\pdfoutput=1

\documentclass[11pt]{article}
\usepackage{naacl2021}
\usepackage{times}
\usepackage{latexsym}
\usepackage{amsmath}
\usepackage{amsfonts}
\usepackage{numprint}

\usepackage{graphicx}
\usepackage{tikz}

\usepackage{mathdots}
\usepackage{cancel}
\usepackage{color}
\usepackage{array}
\usepackage{multirow}
\usepackage{amssymb}
\usepackage{gensymb}
\usepackage{tabularx}
\usepackage{booktabs}
\usepackage{todonotes}
\usetikzlibrary{fadings}
\usetikzlibrary{patterns}
\usetikzlibrary{shadows.blur}
\usetikzlibrary{shapes}

\newcommand{\ignore}[1]{}

\newenvironment{itemizesquish}[2]{\begin{list}{\labelitemi}{\setlength{\itemsep}{#1}\setlength{\labelwidth}{#2}\setlength{\leftmargin}{\labelwidth}\addtolength{\leftmargin}{\labelsep}}}{\end{list}}

\usepackage{microtype}

\DeclareMathOperator*{\argmax}{argmax}

\title{Unsupervised Extractive Summarization by Human Memory Simulation}
\author{Ronald Cardenas$^\clubsuit$  \enspace Matthias Galle$^\spadesuit$ \enspace Shay B. Cohen$^\clubsuit$\\
  $^\clubsuit$ School of Informatics, University of Edinburgh \\
  $^\spadesuit$ Naver Labs Europe \\
  \texttt{ronald.cardenas@ed.ac.uk}  \\
  \texttt{matthias.galle@naverlabs.com} \\
  \texttt{scohen@inf.ed.ac.uk}
  }
  
\date{}

\begin{document}
\maketitle
\begin{abstract}

Summarization systems face the core challenge of identifying and selecting important information.
In this paper, we tackle the problem of content selection in unsupervised extractive summarization of long, structured documents.
We introduce a wide range of heuristics that leverage cognitive representations of content units and how these are 
retained or forgotten in human memory.
We find that properties of these representations of human memory can be exploited to capture relevance of content units in scientific articles.
Experiments show that our proposed heuristics are effective at leveraging cognitive structures and the organization of the document (i.e.\ sections of an article), and automatic and human evaluations provide strong evidence that these heuristics
extract more summary-worthy content units.

% , compared to heuristics that use frequency counts.

% Finally, we revisit the necessity of including a supervised baseline for reference,
% not only to serve as an upperbound in performance but also to put in perspective the practical usability of an unsupervised model.

%We explore a wide range of strategies to exploit properties of these memory representations and identify relevant content units.
%In order to ensure a fair comparison of models, we produce summaries of comparable number of tokens.
%We do so by formulating sentence selection as an optimization problem constrained to a budget.
%We also propose a neural architecture that leverages inductive bias from working memory structures.
%Under low-resource conditions, we find that our model outperforms a strong unsupervised baseline tuned on the same training samples, 
%casting light on the necessity of a truly unsupervised approach.

\end{abstract}

\section{Introduction}

Automatic summarization is the task of presenting a user with a short, computer-generated text that retains the important information from a single or a collection of documents.
The produced summary is expected to be coherent and contain information that is relevant, non-redundant, and informative with respect to the total of information consumed.
Among the many variants of summarization tasks, the task of \textit{extractive} single-document summarization consists of retrieving contiguous chunks of text --usually complete sentences-- from a single source document.
Despite the advance attained in the last few years in the area, the problem of content selection remains an open challenge \cite{narayan2018ranking,kedzie2018content}.

In this work, we tackle the problem of unsupervised extractive summarization of single documents, taking special interest in the problem of content selection.
To this end, we resort to how content is explicitly represented and organized in short-term memory, as modeled by the cognitive theory of human reading comprehension proposed by \citet{kintsch1978toward}, henceforth, KvD. We experiment on long, structured documents, using the body of scientific articles 
%in the biomedical domain 
as documents and their abstracts as summary references.

According to KvD, human working memory --a type of short-term memory-- is composed of content units connected through semantic coherence.
Then, the complete structure, shaped as a tree, conveys a coherent representation of the most relevant information read so far.
We leverage the properties of nodes in these structures, called memory trees, in order to quantify relevance of content units in a sentence.
For example, the relevance of a content unit can be signaled by its position in the memory tree, i.e.\ the closer to the root the more relevant it is.
In summary, our contributions are the following:\footnote{Code available at:\\
\url{https://github.com/ronaldahmed/unsupervised-extr-summarization-kvd}}
\begin{itemizesquish}{-0.3em}{0.5em}
\item We introduce a principled method to explore a varied range of completely unsupervised heuristics for extractive summarization of long documents.
Most notably, the method allows to leverage structures of content units obtained from a cognitive model of reading comprehension.

\item We argue that properties of working memory trees can be exploited to capture relevance of content units in the text and 
improve upon content selection.

\item We formulate the problem of sentence selection as an optimization problem constrained to a budget of number of tokens.
The resulting summaries present less variability in length and are closer to the budget, hence assuring a fairer comparison among models in terms of ROUGE score \cite{lin2004rouge}.

% \item We reflect on the need for comparison of unsupervised models against supervised baselines by examining a recently proposed summarizer \cite{zheng2019pacsum} that requires 
% a small set of gold summary to finetune on.
% We find that a simple sequence classifier based on \textsc{Bert} \cite{devlin2019bert} outperforms this model.
% Furthermore, we propose a supervised model that learns cognitive inductive bias from KvD memory trees, improving slightly upon the baselines.

\end{itemizesquish}

%We investigate the case of long, structured documents, namely scientific articles in the biomedical domain.
%Our experiments show that heuristics leveraging memory tree properties are able to rank relevant content units.
%When compared against content units extracted by an extractive oracle, we find that KvD-aware heuristics extract significantly more summary-worthy content units.

\section{Related Work}

The application of cognitive theories of reading comprehension in summarization tasks has received increased attention in the last few years \cite{zhang2016coherent,lloret2012text}.
Early work by \citet{fang2014summariser} introduced an effective computational implementation of the KvD theory and proposed an extractive summarizer based on 
greedy sentence ranking. Later on, \citet{fang2019proposition} incorporated a generation module in order to produce abstractive summaries from top-ranked content units.
This work builds upon this line of research in two aspects.

First, we introduce a generalized way of aggregating scores of content units, covering a varied range of heuristics beyond count aggregation \cite{fang2014summariser}, 
allowing the proposed models to exploit properties of KvD memory trees as well as the organization of the source document, for instance, by restricting aggregation inside sections in a scientific article.
Second, we propose strategies to select sentences under a budget of number of tokens, providing an appropriate scenario for comparison of systems.
Previous work \cite{narayan2018document,schumann2020discrete} has pointed out the --rather ignored nowadays-- importance of comparing models that produce summaries of similar number of tokens,
highlighting the sensitivity to summary length of popular metrics such as ROUGE.
Our proposed sentence selection strategies directly tackle this area.

Similar to our approach, previous work on unsupervised extractive summarization modeled a document as a graph with sentences or phrases as nodes, ranking them according to their relevance in the network \cite{zheng2019pacsum,mihalcea2004textrank}.
However, these models build a single graph representing the entire information body at once.
In contrast, our system operates incrementally by consuming one sentence at a time, simulating the reading process of a human.
This process, modeled by KvD, takes into consideration the capacity limitations of the human memory.
Hence, at any given time during the reading process, only the most important information will be found in the memory structure instead of the entire network of content units.

% TODO 

\section{The KvD model of Human Memory}
\label{section:kvd-model}

The KvD theory \cite{kintsch1978toward} aims to explain how information is represented and organized in human working memory, a type of temporal storage equipped with mechanisms for updating and reinforcing information.
This theory states that information in the working memory is organized in two levels of semantic structure, micro-structure and macro-structure.
Micro-structure models local coherence and cohesiveness of the text, whilst macro-structure models the general organization of the entire document. 

In this work, we consider only the structures represented at the micro level.
At this level, content units are modeled as propositions of the form \texttt{predicate(arg0,arg1,...)}, and memory is modeled as a tree of propositions, the \textit{memory tree}.
This memory structure presents many convenient properties relevant to the task of summarization.
First, KvD states that the root of the tree should contain information central to the argumentation represented in the working memory; hence, the root is deemed as the most relevant proposition read so far, and the more relevant a proposition is, the closer to the root it will be.
Second, tree branches can be seen as ramifications of the current topic, each branch adding more specialized content as it grows deeper.

%We hypothesize that the relevance of a proposition can be signaled by its properties as a node in a memory tree.

According to KvD, reading is carried out iteratively in \textit{memory cycles}.
In each cycle, only one new sentence is loaded to the working memory. Then, propositions are extracted and added to the current memory tree.
The limits of memory capacity is modeled as a hard constraint in the number of propositions that will be preserved for the next cycle.
Hence, the tree is pruned and some propositions are dropped or \textit{forgotten}.
However, if nodes cannot be attached to the tree in upcoming cycles, these forgotten nodes can be recalled and added to the tree, serving as linking ideas that preserve the local coherence represented in the current tree.\footnote{It is worth noting that \citet{kintsch1978toward} did not specify how many nodes can be recalled at a single time, however recent implementations \cite{fang2019proposition} limit this number to at most 2.}

We now illustrate with an example how content units are captured, forgotten, and recalled during a KvD simulation of reading.

\subsection{Simulation Example}

Consider the first three sentences of the introduction section of a biomedical article, along with its abstract, showed in Table~\ref{table:example-kvd}.
At the beginning of each cycle, propositions are extracted from the incoming sentence and connected to the existing memory tree, obtaining trees (1a), (2a), and (3a).
Two nodes will be connected if they share an argument. For instance, node 5 and 6 share the argument \texttt{antioxidants}.
Then, the most relevant nodes are selected using KvD's \textit{leading edge} strategy, a strategy that aims to keep the most general and most recent nodes.
In this example, we set the memory limit to 5 selected propositions per cycle. The rest are pruned, obtaining trees (1b), (2b), and (3b).
These pruned trees constitute the final product of each cycle and will be used for our content selection experiments.

Let us now analyze what kind of information KvD preserves, forgets or recalls and how it is done.
Note that in cycle 1 the selected root is node $4$, a proposition containing the main verb of the sentence.
However, the root is changed at the beggining of cycle 2 to node $7$ (\texttt{nonenzimatic antioxidants}, \texttt{\#7}), reflecting the change in focus.
Note that \texttt{\#7} is the only proposition mentioned in both sentences, hence serving as link to insert the other nodes.
Now, the tree showcases clearly two ramifications of the current topic, namely that \textit{`\texttt{\#7} control a specific kind of molecules'} and \textit{`deficit of \texttt{\#7} causes certain condition'}.
After selection takes place, the tree is rotated so that the new root reflects the main topic amongst the preserved nodes, resulting in node $10$ being root of (2b).

At the beginning of cycle 3, the newly extracted nodes (14 - 17) cannot be attached to the current tree because the linking node, \texttt{\#8}, was prunned in the previous cycle.
Therefore, information in proposition $8$ is \textit{recalled} and re-attached to the tree, showed as a squared node in tree (3a) and (3b).
Then, the selection strategy is applied and the resulting tree is rebalanced, obtaining (3b).

After analyzing how trees are shaped in each cycle, it is important to point out their importance for the task of extractive summarization.
A sentence ranking system that relies on proposition scoring would first need to capture the right propositions.
Let us look at the first sentence of the gold summary (bottom row in Table~\ref{table:example-kvd}).
On the one hand, many propositions captured by memory trees ($7$, $8$, $12$, $13$, and $15$) appear verbatim in this sentence, although sometimes only partially (e.g.\ $7$ and $15$).
The capture of proposition $8$ in cycle 3 highlights the importance of the recall mechanism in KvD to bring back relevant information.
On the other hand, fine-grained information, relevant to the summary, might also be lost, for instance node $14$ in which a crucial property of a noun is not captured (`\textit{pulmonary}').

\begin{table*}[!h]

\centering
\footnotesize
% \begin{tabular}{|p{0.10\textwidth}|p{0.59\textwidth}|p{0.31\textwidth}|}        
\begin{tabular}{p{7.5cm}p{5cm}}
\hline
\multicolumn{2}{p{15cm}}{\textbf{Cycle 1}\newline
in healthy people , reactive oxidant species are controlled by a number of enzymatic and nonenzymatic antioxidants .} \\

\begin{tabular}[t]{p{9cm}}
\\
  \texttt{1: healthy(people)} \\
  \texttt{2: reactive(species)} \\
  \texttt{3: oxidant(species)} \\
  \texttt{4: are controlled(antioxidants,species, \newline in:people)} \\
  \texttt{5: of(a number,antioxidants)} \\
  \texttt{6: enzymatic(antioxidants)} \\
  \texttt{7: nonenzimatic(antioxidants)}
\end{tabular}

& \begin{tabular}[t]{c}
\tikzset{every picture/.style={line width=0.75pt}} %set default line width to 0.75pt        

\begin{tikzpicture}[x=0.75pt,y=0.75pt,yscale=-1,xscale=1,baseline=0pt]
%uncomment if require: \path (0,188); %set diagram left start at 0, and has height of 188

%Straight Lines [id:da3176831738866588] 
\draw  [dash pattern={on 0.84pt off 2.51pt}]  (25,20) -- (48.33,20) ;
%Straight Lines [id:da21811443436626066] 
\draw    (25,20) -- (49.33,39.67) ;
%Straight Lines [id:da8389522237869071] 
\draw    (25,20) -- (50.33,58.67) ;
%Straight Lines [id:da4884644632378745] 
\draw    (25,20) -- (50.33,79.67) ;
%Straight Lines [id:da4982187608155566] 
\draw  [dash pattern={on 0.84pt off 2.51pt}]  (63.33,81) -- (89.33,81) ;
%Straight Lines [id:da484190913242432] 
\draw    (63.33,81) -- (90.33,99.67) ;
%Straight Lines [id:da815891204759543] 
\draw    (125,20) -- (149.29,20) ;
%Straight Lines [id:da48737986161156743] 
\draw    (125,20) -- (149.29,40.43) ;
%Straight Lines [id:da8303323353360181] 
\draw    (125,20) -- (149.29,60.43) ;
%Straight Lines [id:da14584720732135792] 
\draw    (163.33,61) -- (179.29,61) ;

% Text Node
\draw (52,33) node [anchor=north west][inner sep=0.75pt]  [align=left] {2};
% Text Node
\draw (52,13) node [anchor=north west][inner sep=0.75pt]  [align=left] {1};
% Text Node
\draw (12,13) node [anchor=north west][inner sep=0.75pt]  [align=left] {4};
% Text Node
\draw (52,53) node [anchor=north west][inner sep=0.75pt]  [align=left] {3};
% Text Node
\draw (52,73) node [anchor=north west][inner sep=0.75pt]  [align=left] {5};
% Text Node
\draw (92,74) node [anchor=north west][inner sep=0.75pt]  [align=left] {6};
% Text Node
\draw (92,93) node [anchor=north west][inner sep=0.75pt]  [align=left] {7};
% Text Node
\draw (152,13) node [anchor=north west][inner sep=0.75pt]  [align=left] {2};
% Text Node
\draw (112,13) node [anchor=north west][inner sep=0.75pt]  [align=left] {4};
% Text Node
\draw (152,33) node [anchor=north west][inner sep=0.75pt]  [align=left] {3};
% Text Node
\draw (152,53) node [anchor=north west][inner sep=0.75pt]  [align=left] {5};
% Text Node
\draw (182,54) node [anchor=north west][inner sep=0.75pt]  [align=left] {7};
% Text Node
\draw (-15,10) node [anchor=north west][inner sep=0.75pt]   [align=left] {(1a)};
% Text Node
\draw (85,10) node [anchor=north west][inner sep=0.75pt]   [align=left] {(1b)};

\end{tikzpicture}
\end{tabular} \\ \\

\multicolumn{2}{p{15cm}}{
\textbf{Cycle 2}\newline
in patients with cystic fibrosis ( cf ) , deficiency of nonenzymatic antioxidants is linked to malabsortion of lipib-soluble vitamins .} \\

\begin{tabular}[t]{p{10cm}}
\\
	\texttt{8: with(patients,cystic fibrosis)} \\
	\texttt{9: BE(cystic fibrosis,cf)} \\
	\texttt{10: of(deficiency, \#7)} \\
	\texttt{11: is linked (malabsortion,\#10, in:\#8)} \\
	\texttt{12: of (malabsortion,vitamins)} \\
	\texttt{13: lipid-soluble(vitamins)}
\end{tabular}

& \begin{tabular}[t]{c}

\tikzset{every picture/.style={line width=0.75pt}} %set default line width to 0.75pt        

\scalebox{0.75}{
\begin{tikzpicture}[x=0.75pt,y=0.75pt,yscale=-1,xscale=1,baseline=0pt]
%uncomment if require: \path (0,220); %set diagram left start at 0, and has height of 220

%Straight Lines [id:da39513801964452] 
\draw  [dash pattern={on 0.84pt off 2.51pt}]  (19.29,20.33) -- (28.33,20.33) ;
%Straight Lines [id:da06340938641066418] 
\draw  [dash pattern={on 0.84pt off 2.51pt}]  (43.33,20.33) -- (59.33,20.33) ;
%Straight Lines [id:da8533743546896975] 
\draw  [dash pattern={on 0.84pt off 2.51pt}]  (43.33,20.33) -- (60.29,39.86) ;
%Straight Lines [id:da030928316747995233] 
\draw    (19.29,20.33) -- (28.29,90.86) ;
%Straight Lines [id:da9779085024375187] 
\draw  [dash pattern={on 0.84pt off 2.51pt}]  (77.33,89.33) -- (87.33,89.33) ;
%Straight Lines [id:da8451382108782117] 
\draw  [dash pattern={on 0.84pt off 2.51pt}]  (105.33,89.33) -- (117.29,89.33) ;
%Straight Lines [id:da5470480042353187] 
\draw    (77.33,89.33) -- (86.33,119.33) ;
%Straight Lines [id:da8913845897165456] 
\draw    (47.33,89.33) -- (59.33,89.33) ;
%Straight Lines [id:da32948629709250254] 
\draw  [dash pattern={on 0.84pt off 2.51pt}]  (43.33,20.33) -- (58.29,58.86) ;
%Straight Lines [id:da5105986118497066] 
\draw    (106.33,118.33) -- (117.29,118.33) ;
%Straight Lines [id:da4718896612409982] 
\draw    (160.33,19.33) -- (172.33,19.33) ;
%Straight Lines [id:da9331645759163611] 
\draw    (187.33,49.33) -- (197.14,49.33) ;
%Straight Lines [id:da6052062916651095] 
\draw    (214.33,49.33) -- (224.14,49.33) ;
%Straight Lines [id:da8264294089411159] 
\draw    (160.33,19.33) -- (170.14,48.86) ;

% Text Node
\draw (31,13) node [anchor=north west][inner sep=0.75pt]  [align=left] {4};
% Text Node
\draw (8,13) node [anchor=north west][inner sep=0.75pt]  [align=left] {7};
% Text Node
\draw (30,82) node [anchor=north west][inner sep=0.75pt]  [align=left] {10};
% Text Node
\draw (62,13) node [anchor=north west][inner sep=0.75pt]  [align=left] {2};
% Text Node
\draw (60,82) node [anchor=north west][inner sep=0.75pt]  [align=left] {11};
% Text Node
\draw (62,53) node [anchor=north west][inner sep=0.75pt]  [align=left] {5};
% Text Node
\draw (62,33) node [anchor=north west][inner sep=0.75pt]  [align=left] {3};
% Text Node
\draw (87,112) node [anchor=north west][inner sep=0.75pt]  [align=left] {12};
% Text Node
\draw (91,82) node [anchor=north west][inner sep=0.75pt]  [align=left] {8};
% Text Node
\draw (117,82) node [anchor=north west][inner sep=0.75pt]  [align=left] {9};
% Text Node
\draw (117,112) node [anchor=north west][inner sep=0.75pt]  [align=left] {13};
% Text Node
\draw (141,13) node [anchor=north west][inner sep=0.75pt]  [align=left] {10};
% Text Node
\draw (172,43) node [anchor=north west][inner sep=0.75pt]  [align=left] {11};
% Text Node
\draw (197,43) node [anchor=north west][inner sep=0.75pt]  [align=left] {12};
% Text Node
\draw (222,43) node [anchor=north west][inner sep=0.75pt]  [align=left] {13};
% Text Node
\draw (173,13) node [anchor=north west][inner sep=0.75pt]  [align=left] {7};
% Text Node
\draw (-20,10) node [anchor=north west][inner sep=0.75pt]   [align=left] {(2a)};
% Text Node
\draw (105,10) node [anchor=north west][inner sep=0.75pt]   [align=left] {(2b)};

\end{tikzpicture}
}
\end{tabular} \\ \\

\multicolumn{2}{p{15cm}}{\textbf{Cycle 3}\newline
furthermore, pulmonary inflammation in cf patients also contributes to depletion of antioxidants .} \\

\begin{tabular}[t]{p{6cm}}
\\
	\texttt{14: pulmonary(inflammation)} \\
	\texttt{15: inflammation(in:\#8)} \\
	\texttt{16: contributes(\#15,to:depletion)} \\
	\texttt{17: of(depletion,antioxidants)}
\end{tabular}
 
& \begin{tabular}[t]{c}

\tikzset{every picture/.style={line width=0.75pt}} %set default line width to 0.75pt        
\scalebox{0.75}{
\begin{tikzpicture}[x=0.75pt,y=0.75pt,yscale=-1,xscale=1,baseline=0pt]
%uncomment if require: \path (0,231); %set diagram left start at 0, and has height of 231

%Straight Lines [id:da14133687569821918] 
\draw  [dash pattern={on 0.84pt off 2.51pt}]  (31.33,20.33) -- (43.33,20.33) ;
%Straight Lines [id:da3329860223159753] 
\draw  [dash pattern={on 0.84pt off 2.51pt}]  (58.33,50.33) -- (68.14,50.33) ;
%Straight Lines [id:da5054060148291082] 
\draw  [dash pattern={on 0.84pt off 2.51pt}]  (85.33,50.33) -- (93.4,50.33) ;
%Straight Lines [id:da7942233230251767] 
\draw    (31.33,20.33) -- (41.14,49.86) ;
%Straight Lines [id:da09318173144056052] 
\draw    (84.33,80.33) -- (92.4,80.2) ;
%Straight Lines [id:da5346985672188882] 
\draw  [dash pattern={on 0.84pt off 2.51pt}]  (111.33,80.33) -- (118.4,80.33) ;
%Straight Lines [id:da3802107416742091] 
\draw    (58.33,50.33) -- (69.4,81.2) ;
%Flowchart: Process [id:dp5672136850890213] 
\draw   (69.4,71.2) -- (84.4,71.2) -- (84.4,90.2) -- (69.4,90.2) -- cycle ;
%Straight Lines [id:da04819390285164182] 
\draw    (111.33,80.33) -- (118.4,110.2) ;
%Straight Lines [id:da9251556160045347] 
\draw  [dash pattern={on 0.84pt off 2.51pt}]  (135.33,110.33) -- (143.4,110.33) ;
%Flowchart: Process [id:dp8895356217877288] 
\draw   (202.4,40.2) -- (217.4,40.2) -- (217.4,59.2) -- (202.4,59.2) -- cycle ;
%Straight Lines [id:da2602940183386626] 
\draw    (190,20) -- (202.14,49.86) ;
%Straight Lines [id:da9940177693459271] 
\draw    (190,20) -- (202,20) ;
%Straight Lines [id:da071149142670486] 
\draw    (218,49) -- (230,49) ;
%Straight Lines [id:da19600627004169935] 
\draw    (247,49) -- (255,49) ;

% Text Node
\draw (12,14) node [anchor=north west][inner sep=0.75pt] [align=left] {10};
% Text Node
\draw (43,44) node [anchor=north west][inner sep=0.75pt]  [align=left] {11};
% Text Node
\draw (68,44) node [anchor=north west][inner sep=0.75pt]  [align=left] {12};
% Text Node
\draw (93,44) node [anchor=north west][inner sep=0.75pt]  [align=left] {13};
% Text Node
\draw (44,14) node [anchor=north west][inner sep=0.75pt]   [align=left] {7};
% Text Node
\draw (72,73) node [anchor=north west][inner sep=0.75pt]  [align=left] {8};
% Text Node
\draw (93,73) node [anchor=north west][inner sep=0.75pt]  [align=left] {15};
% Text Node
\draw (118,73) node [anchor=north west][inner sep=0.75pt]  [align=left] {14};
% Text Node
\draw (118,103) node [anchor=north west][inner sep=0.75pt]  [align=left] {16};
% Text Node
\draw (144,103) node [anchor=north west][inner sep=0.75pt] [align=left] {17};
% Text Node
\draw (172,13) node [anchor=north west][inner sep=0.75pt]  [align=left] {11};
% Text Node
\draw (202,13) node [anchor=north west][inner sep=0.75pt]  [align=left] {10};
% Text Node
\draw (204.4,43.2) node [anchor=north west][inner sep=0.75pt]  [align=left] {8};
% Text Node
\draw (230,42) node [anchor=north west][inner sep=0.75pt]  [align=left] {15};
% Text Node
\draw (255,42) node [anchor=north west][inner sep=0.75pt]  [align=left] {16};
% Text Node
\draw (-20,10) node [anchor=north west][inner sep=0.75pt]   [align=left] {(3a)};
% Text Node
\draw (130,10) node [anchor=north west][inner sep=0.75pt]   [align=left] {(3b)};

\end{tikzpicture}
}
\end{tabular} \\ \\
\hline 

\multicolumn{2}{p{15cm}}{\textbf{Gold Summary}\newline patients with cystic fibrosis ( cf ) show decreased plasma concentrations of antioxidants due to malabsorption of lipid soluble vitamins and consumption by chronic pulmonary inflammation .\newline
carotene is a major source of retinol and therefore is of particular significance in cf .\newline
... } \\ \hline
\end{tabular}

\caption{Simulation of KvD reading during three cycles. Each row shows the sentence consumed (top),
	the propositions extracted (left), and memory trees before (1a, 2a, 3a) and after (1b, 2b, 3b) applying a memory constraint of 5 nodes per cycle.
	Argument \texttt{\#N} means that proposition $N$ is used as argument.
	Squared nodes are recalled propositions.
	Solid lines connect nodes selected to keep in memory, and dotted lines connect nodes to be pruned or forgotten. }
\label{table:example-kvd}
\end{table*}

\subsection{Reproduction Probability}
In each cycle, a proposition can either be selected to stay in the working memory tree or removed from it and sent to long-term memory.
At the end of the simulation, a previously removed proposition can still be used in the summary if it was relevant enough.
KvD captures this relevancy through the \textit{reproduction probability} parameter, $\rho$, which expresses the probability that in a single cycle certain proposition is stored in long-term memory and later retrieved during summarization.

Hence, if a proposition participates in $k$ cycles, each time with probability $\rho$ of being removed, its reproduction probability at the end of the simulation will be defined as
\begin{equation}
\label{eq:rprob}
rp_k = 1 - (1-\rho)^k.
\end{equation}

% \subsection{KvD and discourse theories}

\section{Summarization using Memory Trees}

We formulate the problem of unsupervised extractive summarization as the task of scoring the sentences in a document followed by a selection step in which an optimal set of sentences is chosen as the summary.

\subsection{Sentence Scoring}
\label{section:sent-sc}
We define the score of a sentence as the sum of the scores of all the propositions found in that sentence $s$, namely
%\begin{align*}
 $sc(s) = \sum_{p \in s} v(p)$,
%\end{align*}
where $v(p)$ is the score of proposition $p$.
In order to calculate $v(p)$, we first score each occurrence of $p$ as a node in memory trees.
Then, all occurrence scores are aggregated into $v(p)$. We propose a variety of heuristics for each step which we now elaborate.

\paragraph{Occurrence scoring.}
We call an \textit{occurrence} of a proposition $p$ during simulation to every instance of $p$ where it appears as a node in a memory tree.
Since a memory cycle can keep a proposition in the tree for the next cycle, there can be many such instances for a certain $p$.

Let $\mathcal{N}(p)$ be the set of occurrences of $p$ as a node in a memory tree during simulation of the entire document.
For each $x \in \mathcal{N}(p)$, the scorer $c(x)$ is defined as one of the following:
\begin{tabular}{l|l}
$c_{cnt}(x) = 1$, & $c_{lvl}(x) = \frac{1}{depth(x)}$, \\
$c_{deg}(x) = degree(x)$, & $c_{sub}(x) = |t_x|$,
\end{tabular}

\noindent where, \textit{depth(x)} is the depth of node $x$ with respect to the tree root; \textit{degree(x)} is the degree of node $x$ in the tree; and $|t_x|$ is the size of the subtree rooted in $x$.

\paragraph{Aggregation of occurrence score.}
Occurrence scores are aggregated depending on whether we consider occurrences in the entire document or occurrences by section, as follows
\begin{align*}
n_{cnt} &= \sum_{x \in \mathcal{N}(p)} c(x), \\
n_{wgt} &= \sum_{y \in Y} \left[ r_y \cdot \left( \sum_{x \in \mathcal{N}_y(p)} c(x) \right) \right], \\
n_{exp} &= \sum_{y \in Y} \left[ \sum_{x \in \mathcal{N}_y(p)} c(x)\right] ^ {r_y},
\end{align*}
\noindent where $\mathcal{N}_y(p)$ is the set of occurrences of $p$ during simulation of section $y \in Y=\{\textit{Introduction, Discussion, Conclusion}\}$,
and $r_y$ is the ratio of sentences in section $y$. For instance, for the \textit{Introduction} section,
\begin{equation*}
r_i= \frac{\textit{Number of sentences in the introduction}}{\textit{Total number of sentences in the document}}
\end{equation*}

\paragraph{Proposition score.}
Finally, the score of a proposition $p$ is defined as
\begin{equation}
v(p) =  1-(1-\rho)^{n(p)},
\label{eq:v-p}
\end{equation}

% which could be interpreted as a generalization to Equation~\ref{eq:rprob}, in which KvD's expression corresponds to the case when $n(p)$ is raw frequencies.

\paragraph{Combined heuristic configuration.}
For ease of notation, a heuristic $v(p)$ with configuration $c_a$ and $n_b$ will be referred to as heuristic \texttt{a-b}.
For instance, heuristic \texttt{Lvl-Exp} refers to a heuristic that combines occurrence scoring by node depth ($c_{lvl}$) and
aggregates the scores by document section as an exponentially weighted sum ($n_{exp}$).

Notice that Equation~\ref{eq:v-p} can be seen as a generalization of KvD's definition of reproduction probability where heuristic \texttt{Cnt-Cnt} is equivalent to Equation~\ref{eq:rprob}.
Moreover, the flexibility in configuration allows to choose to exploit either the shape of memory trees or to exploit the structure of a document, or both at the same time.
First, configurations using $c_{lvl}$, $c_{deg}$, and $c_{sub}$ do exploit the shape and configuration of the trees, whereas
those using $c_{cnt}$ do not.
Second, configurations using $n_{wgt}$ and $n_{exp}$ leverage the fact that the document is divided in sections, whereas configurations using $n_{cnt}$ do not.

% The combined space of configurations includes the following general cases.
% First, heuristic $KvD_{cnt-cnt}$ is equivalent to counting the number of occurrences of $p$, $|\mathcal{N}(p)|$.
% Second, heuristic \texttt{raw-raw-rprob} is equivalent to Equation~\ref{eq:rprob} and configurations of the form \texttt{*-*-rprob}
% can be seen as generalizations of KvD's definition of reproduction probability.

\subsection{Sentence Selection}

Previous work has pointed out that ROUGE score is sensitive to the length of the summary and summarization models should only be compared against each other if they produce summaries of similar length \cite{narayan2018document,schumann2020discrete}.
For this reason, we choose to extract summaries according to a budget of tokens instead of picking a fixed number of sentences regardless of their length as is normally reported in the literature.

Given a document $\mathcal{D}=\langle s_0,s_1,...,s_N\rangle$, heuristic sentence scorer $sc:s_i \to \mathbb{R}$, and a budget of tokens $W$, the summary is extracted as follows.

\paragraph{\textsc{Greedy}.}
The summary is defined as the top scored sentences which total length is less than or equal to $W$.

\paragraph{\textsc{Shorter}.}

We adapt the 0-1 knapsack problem to the sentence selection problem.
The objective is to maximize the total score of selected sentences while complying with a budget $W$.
Each sentence contributes to the budget with its length in number of tokens.
Formally, the optimal summary $\mathcal{S}$ is defined as
\begin{equation*}
\label{eq:shorter}
\mathcal{S} = \argmax_{\hat{S}} \sum_{s_j \in \hat{S}} sc(s_j),\, \textit{s.t.}\sum_{s_j \in \hat{S}} |s_j| \leq W,
\end{equation*}

\paragraph{\textsc{Closest}.}
This strategy extends \textsc{Shorter} by relaxing the budget constraint and allowing longer summaries to be considered.
A longer summary will be preferred over the previous (budget-abiding) best if its score is higher and its length is closer to the budget.
Formally, $\mathcal{S}$ is defined as
\begin{equation*}
\label{eq:closest}
\mathcal{S} = \argmax_{\hat{S}} \sum_{s_j \in \hat{S}} sc(s_j) \textit{ s.t. }\, |\hat{S}|-W < | |\tilde{S}|-W|,
\end{equation*}

\noindent where $|\hat{S}|$ is the number of tokens in candidate summary $\hat{S}$ and $\tilde{S}$ is the previous best candidate that met the budget constraint.

\section{Experimental Setup}

We investigate whether the organization of content units, as modeled by trees of propositions, is an effective signal to rank sentences and 
obtain sensible extractive summaries.
To this end, we test all possible combinations of the occurrence scoring and aggregation strategies presented in Sect.~\ref{section:sent-sc}.
For sentence scoring, we use ratios of sentences per section $r_i=0.33$ for \textit{Introduction}, $r_d=0.53$ for \textit{Discussion}, and $r_c=0.14$ for \textit{Conclusion}.
We use reproduction probability $\rho=0.3$ and memory limit constraint $M=\{5,20,50,100\}$.
All models in our experiments (including supervised and semi-supervised baselines) operate under the \textsc{Closest} selection strategy with a budget $W$ of 205 tokens--the average gold summary length in the training set.

\subsection{Dataset}

We use the PubMed dataset collected by \citet{cohan2018discourse}, composed of scientific articles in the biomedical domain with their abstracts as reference summaries.
We only consider the Introduction, Discussion, and Conclusion sections in each article, as preliminary experiments showed that most information needed to summarize the document is found there.
After filtering out articles without any of these sections, we end up with \numprint{104814} articles in the training set, \numprint{5344} in the validation set, and \numprint{6025} in the test set.
We randomly select \numprint{1000} instances from the training set under a uniform distribution in order to finetune an unsupervised baseline and train a supervised one.
%This subsampled data is also used for one of our unsupervised baselines which requires a small set of gold summaries to tune parameters.

We report ROUGE recall scores \cite{lin2004rouge} instead of F$_1$ scores.
While reporting ROUGE F$_1$ scores is common with abstractive summarization, \newcite{narayan2018document} and \newcite{schumann2020discrete} pointed out that F$_1$ score is significantly sensitive to summary length and that recall values are more appropriate for extractive summarization when summary lengths are similar.

\subsection{KvD Simulator}

We use the KvD implementation proposed by \newcite{fang2019proposition} which produces micro trees of linguistic propositions extracted from a document.
The resulting trees, one per sentence, comply with the memory limit constraint proposed by KvD, i.e.\ all trees have at most $M$ nodes.
First, the simulator extracts dependency trees and performs coherence resolution on a document using Stanford CoreNLP v3.9.2 \cite{stanfordcorenlp}.
Second, the simulator extracts propositions from the complete document and calculates semantic relatedness scores between them.
Finally, reading is simulated as described in Sect.~\ref{section:kvd-model}.

It is worth mentioning that we reset the KvD simulator at the beginning of each section of the article in order to generate memory trees that reflect only the argumentation of the current section but still having access to the complete set of propositions in the document, in case a content unit is referenced back.
In this way, we force the KvD simulator to produce memory trees with nodes only relevant to the current section.

\subsection{Extractive Oracle Under Constraints}
It is common practice to extract sentences from the source document to serve as an oracle summary for supervised extractive systems.
Previous work has applied a greedy approach by extracting the subset of sentences that maximizes the ROUGE score, typically the sum of ROUGE-1 and ROUGE-2 F$_1$ values.
This approach starts with an empty set and adds one sentence at a time, stopping when the maximum number of sentences is reached.

We adapt our sentence selection strategies under budget constraints to obtain oracle sentences.
The score of an oracle summary is the sum of ROUGE-1 and ROUGE-2 recall values calculated with respect to the gold summary.

\subsection{Baselines}
We report the following baselines.
\begin{itemizesquish}{-0.3em}{0.5em}
\item \textsc{Lead}. The first sentences until budget is reached.
\item \textsc{Longest}. Pick sentences in descending order of length in tokens until budget is reached.
\item \textsc{Random}. The score of each sentence is its probability, drawn from a uniform distribution. Then the selection strategy is applied.
\item \textsc{Random-Wgt}. The score of each sentence is its probability, proportional to the ratio of the section it belongs to.
\item \textsc{NoTree}. Heuristic configuration that counts proposition occurrences in the source document instead of occurrences in memory trees.
\item \textsc{PacSum}. Unsupervised model \cite{zheng2019pacsum} that models sentences as nodes in a graph, ranking them based on node centrality. We employ the tf-idf scorer, labeled as \textsc{PacSum(tfi-df)} in \citet{zheng2019pacsum}.
We use two configurations of this model in our experiments. The first one, labeled simply as \textsc{PacSum}, uses the default hyper-parameters reported by \citet{zheng2019pacsum}. The second one uses hyper-parameters fine-tuned over a sample of \numprint{1000} articles from the training set, and we call it \textsc{PacSum-FT}.

\end{itemizesquish}

\paragraph{Supervised baseline.}
In addition to the aforementioned unsupervised baselines, we compare our models against a supervised baseline based on SciBert \cite{beltagy-etal-2019-scibert} and using the pretrained models served by HuggingFace.\footnote{\url{https://huggingface.co/allenai/scibert_scivocab_uncased}}
We add a linear classifier layer on top of the transformer model and fine-tune it over the same subset used to fine-tune \textsc{PacSum-FT}.
In a similar fashion to \citet{cohan2019pretrained}, we consume each document in chunks of fixed numbers of sentences.
Optimization details can be found in Appendix A. %~\ref{app:scibert}.
We refer to this baseline as \textsc{SciBert}.

\subsection{Analysis of Selection Strategy}

We performed preliminary experiments in order to investigate the properties of the proposed selection strategies and determine the most appropriate one.
Intuitively, the closer the output summaries are in length the fairer the comparison among systems.
Therefore, it is desirable that the distribution of summary length values in terms of number of tokens exhibits as low a standard deviation as possible.
Additionally, it is desirable for the mean of length values to be the closest to the budget as possible as to minimize the discrepancy in summary length between gold and predicted summaries.

We explore the entire heuristic configuration space under memory limit values $M$ of 5 and 100 and a budget $W$ of 205 tokens.
For each selection strategy, we analyze the mean and standard deviation of the distribution of summary length values, predicted over the validation set.

For strategy \textsc{Greedy}, the average mean over all heuristics was 179.69 tokens and the average standard deviation, 28.01.
For strategy \textsc{Shorter}, the average mean was 202.20 tokens and the average standard deviation, 12.66.
Finally, for \textsc{Closest}, these numbers where 203.95 and 12.28.
% gold valid set: 205.6366, 77.8678

From these results, we observe that, as expected, a greedy approach to sentence selection does much worse than a combinatorial optimization approach.
%Summaries extracted by \textsc{greedy} were farthest from the budget objective and showed the greatest standard deviation in terms of summary length.
%In contrast, summaries length values under \textsc{Closest} showed the lowest standard deviation and were closest to the budget.
As consequence, we use strategy \textsc{Closest} for the rest of experiments.

\subsection{Extraction per Section}
We investigate whether a model extracts sentences from a section in the document in a similar fashion as the oracle extractor does.
Let $\hat{q}_y$ be the proportion of sentences in candidate summary $\mathcal{S}$ that belong to section $y$,
and let $q_y$ be the equivalent proportion calculated from the oracle summary.

We define metric $q_{\textit{diff}}$ as the divergence of $\hat{q}_y$  w.r.t $q_y$, summed over all sections, as follows
\begin{align}
q_{\textit{diff}} = |q_i-\hat{q}_i| + |q_d-\hat{q}_d| + |q_c-\hat{q}_c|,
\end{align}
where $q_i$, $q_d$, and $q_c$ are proportions for sections \textit{Introduction}, \textit{Discussion}, and \textit{Conclusion}.

Intuitively, it is desirable that $q_{\textit{diff}}$ is as  low as possible, meaning that a summarizer chooses sentences from sections in a similar way as the oracle extractor does.

% Add range of value / worst case scenarios -> intuition about how to interpret this number

%% THIS WHOLE SECTION WAS MOVED DIRECTLY TO THE EXPERIMENTS
%\subsection{Capturing relevant propositions}
%We quantify how many propositions extracted by a model are also present in the extractive oracle summary.
%Let $\mathcal{P}$ be the set of propositions present in oracle summary of document $\mathcal{D}$, and 
%let $\hat{\mathcal{P}}$ be the set of propositions in candidate summary $\mathcal{S}$.
%We define recall (R) and precision (P) as follows
%$R = |\mathcal{P} \cap \hat{\mathcal{P}}|/|\mathcal{P}|$,
%$P = |\mathcal{P} \cap \hat{\mathcal{P}}|/|\hat{\mathcal{P}}|$.
%A higher value of $R$ means that more summary-worthy content is being captured. 
%Along with recall and precision, 
%We also report the F$_1$ score.

%\begin{align*}
%R &= \frac{|\mathcal{P} \cap %\hat{\mathcal{P}}|}{|\mathcal{P}|},
%P = \frac{|\mathcal{P} \cap %\hat{\mathcal{P}}|}{|\hat{\mathcal{P}}|}.
%\end{align*}

\subsection{Human Evaluation}
Additionally, we elicit human judgement in order to evaluate the degree to which our heuristic systems capture key content in a scientific article.
For this we employ a question-answering (QA) paradigm \cite{clarke2010discourse,narayan2018ranking,narayan2019article} with Cloze style queries instead of factoid questions \cite{hermann2015teaching}.
Queries are constructed by replacing one factual detail from the reference summary.
Human subjects are presented with a system summary and a query, and are asked to provide the missing piece of information.

We evaluated heuristic \texttt{Sub-Exp} for tree size $20$, as this heuristic had the highest sum of \textsc{Rouge-1} and \textsc{Rouge-2} scores. As baseline, we evaluate system \textsc{NoTree}, and as control we evaluate \textsc{Oracle}. 
Comparing against \textsc{Oracle} gives us an upper-bound as to how much information can be captured in the optimal scenario.
For completeness, we also include \textsc{PacSum} in our evaluation.

We randomly sampled 50 documents from the test set and manually constructed three queries per document, blurring only one piece of information per query. Each document-system-query combination was answered by three participants through the Amazon Mechanical Turk platform, a total of 600 task items.
We deployed the task items in batches (one system-query combination at a time) to ensure that any single participant is not exposed to system summaries of the same document or to queries built from the same reference summary.
We use the scoring strategy proposed by \citet{clarke2010discourse}, scoring a correct answer (i.e.\ exact string match) with score $1.0$, a partially correct answer (i.e.\ partial string match) with $0.5$, and $0.0$ otherwise.

\section{Results and Discussion}

\subsection{Content Selection at the Sentence Level}

We start by analyzing the performance of our heuristics at selecting relevant sentences from the correct document sections.
In Table~\ref{table:gen} we observe that the organization of information in the dataset articles poses a challenge for trivial baselines.
For instance, \textsc{Lead} does worse than randomly picking sentences (e.g.\ \textsc{Random} and \textsc{Random-wgt}).
Note also that \textsc{Longest} performs poorly after our \textsc{Closest} selection strategy forces output summaries to be close to the budget.

Note that all heuristics perform better than the heuristic baseline \textsc{NoTree} which ranks propositions according to their frequency in the document.
Table~\ref{table:gen} also shows the best and worst heuristic configuration per memory limit, chosen from results in the validation set.
It is worth noting that for every $M$ setup, the worse heuristic belongs to a class that only uses node frequencies in the memory trees and not properties of the tree itself.
In contrast, all of the best heuristics belong to a class that scores occurrences by exploiting the subtree size ($c_{sub}$) or depth of the node in the tree ($c_{lvl}$).
In terms of ROUGE score, we find that a memory limit of 5 during KvD simulation is most effective compared to larger memory buffer sizes.
We hypothesize that a smaller memory tree forces the simulator to keep only the most relevant nodes at that moment.

\begin{table}[t]
\centering
\footnotesize
\begin{tabular}{|l|c|c|c|c|c|c|c|}
\hline
Model         & M & R1    & R2    & RL  & $q_{\textit{diff}}$ \\ \hline
\texttt{Sub-Cnt} & 5 &         \textbf{44.10}  & 14.50  & 39.39    & \textbf{7.79}  \\
\texttt{Cnt-Wgt}   & 5 &           43.35 & 13.65 & 38.55    & 17.71 \\ \hline
\texttt{Sub-Exp} & 20 &          44.00 & \textbf{14.70}  & \textbf{39.40}  & 9.97  \\
\texttt{Cnt-Wgt} & 20 &          42.90 & 13.44 & 38.18     & 18.27 \\ \hline
\texttt{Lvl-Exp} & 50 &          43.51 & 13.99 & 38.81     & 11.81 \\
\texttt{Cnt-Cnt} & 50 &          42.75 & 13.30 & 38.07     & 15.33 \\ \hline
\texttt{lvl-Exp} & 100 &         43.20 & 13.46 & 38.48     & 29.06 \\ 
\texttt{Cnt-Cnt} & 100 &         42.72 & 13.18 & 37.99     & 29.72 \\ \hline
\hline
\multicolumn{2}{|l|}{\textsc{Oracle}}                   & 60.08 & 28.74 & 54.46     &  0 \\
\multicolumn{2}{|l|}{\textsc{Lead}}                     & 41.12 & 13.36 & 36.72     & 83.83 \\
\multicolumn{2}{|l|}{\textsc{Longest}}                  & 41.35 & 12.25 & 35.18     & 10.16 \\
\multicolumn{2}{|l|}{\textsc{Rnd}}                    & 42.91 & 13.06 & 38.04     & 14.28 \\
\multicolumn{2}{|l|}{\textsc{Rnd-Wgt}}               & 42.60 & 12.71 & 37.73     & 29.18 \\
\multicolumn{2}{|l|}{\textsc{NoTree}}                          & 43.20 & 13.27 & 38.51     & 5.93  \\
\multicolumn{2}{|l|}{\textsc{PacSum}}                   & 37.86  & 11.71  & 32.99      & 16.40  \\ \hline
\hline
\multicolumn{2}{|l|}{\textsc{SciBert}  }                          & 47.16                  & 17.37                  & 42.88                                                       & 4.95                       \\
\multicolumn{2}{|l|}{\textsc{PacSum-FT} }                    & 45.81  & 16.36  & 41.04        & 43.34  \\
\hline
\end{tabular}
\caption{Performance in terms of ROUGE recall score. For heuristic rows under a single memory limit $M$, 
the best and worst model are reported at the top and at the bottom, respectively. $|\mathcal{S}|_{avg}$ is the mean summary length in number of tokens. \textsc{PacSum} uses the default hyper-parameters reported by \citet{zheng2019pacsum}. }
\label{table:gen}
\end{table}

We also observe that a smaller memory tree helps the heuristic to select sentences from the right section of the document, as signaled by lower $q_{\textit{diff}}$ values.
In larger memory trees, more propositions get to accumulate score during simulation, hence making longer sentences obtain higher scores.
This can be noted by an increasing average summary length value as the size of tree increases. %In almost all cases, summaries are of average length of $200 \plusminus 7$ tokens.

% picking sentences from the right section
%For reference, the average proportion of sentences in oracle summaries that were extracted from an specific section are
%$q_i=0.33$, $q_d=0.53$, and $q_c=0.14$.

Consider now the supervised upper-bound for this task, \textsc{SciBert}, and the fine-tuned \textsc{PacSum-FT}.
This last one outperforms our best heuristic but still falls behind \textsc{SciBert} by almost one point.
Note however that when using default parameters, we observe a dramatic drop in scores, partially explained by significantly shorter summaries produced by \textsc{PacSum}.
Additionally, it can be observed that both configurations of PacSum struggle to select content from the right document sections, as signaled by the high values of $q_{\textit{diff}}$,
in contrast to \textsc{SciBert}.
This result casts light on the necessity of comparing against a supervised baseline, especially against an unsupervised model that is fine-tuned on gold-standard data, such as \textsc{PacSum}.

% In summary, when presented with small amount of gold reference summaries, a simple supervised model outperforms all the analyzed unsupervised models, including one that was fine-tuned over the same gold data.

\paragraph{Human evaluation.} Our Cloze QA evaluation revealed participants are able to answer a query $89.16\%$ of the time after reading the extractive oracle summary.
When presented with output from \textsc{Sub-exp-rprob}, \textsc{Raw-Doc}, and \textsc{PacSum}, this percentage is $78.0\%$, $73.66\%$, and $72.0\%$, respectively.

% We tested pairwise statistical significance between systems using a one-way ANOVA with posthoc Tukey HSD tests under $p<0.01$. We found that \textsc{Oracle} is significantly different from \textsc{Raw-Doc} and \textsc{PacSum}, all other differences being non-significant.
% On one hand, this could mean that our best heuristic does a great job and it's indistinguishable from the oracle.

%\roncomment{do we want an example here? Check XSum paper, figure 4.}
% https://arxiv.org/pdf/1907.08722.pdf

\subsection{Content Selection at the Proposition Level}

We now analize how many relevant propositions are captured by heuristics, comparing them through precision and recall with their presence or not in a sentence extracted by the oracle.
We compare heuristics using occurrence scorers that exploit a tree property against those who use only frequency, for aggregation strategy $n_{exp}$ and tree size $M=20$, as showed in Table~\ref{table:cnt-sel}. For completeness, we also include baseline \textsc{NoTree}.

We observe that among all tree properties analyzed, using the depth of a node in the tree seems to be most beneficial in terms of F1 score.
Closely behind are found heuristics using the size of the subtree and the node degree.
In contrast, heuristics using only frequency (\texttt{Cnt-Exp} and \textsc{NoTree}) seem to capture less oracle propositions, although they do not fall far away behind.

It is also worth noting that the proportion of oracle propositions captured by the heuristics is low, around 30\%.
Preliminary experiments showed that, even though larger memory limit setups capture a larger number of oracle propositions (around 70\% for $M=100$), 
more noise is also scored higher, hence making sentence selection harder.

\begin{table}[h]
\centering
\begin{tabular}{|l|c|c|c|}
\hline
Heuristic     & R(\%)     & P(\%)     & F1(\%)    \\ \hline
\texttt{Sub-Exp} & 28.83 & \textbf{31.12} & 29.79 \\
\texttt{Lvl-Exp} & \textbf{28.84} & 31.11 & \textbf{29.80} \\
\texttt{Deg-Exp} & 28.80 & 31.08 & 29.76 \\
\texttt{Cnt-Exp} & 28.68 & 30.93 & 29.62 \\
\textsc{NoTree}   & 26.13 & 28.52 & 27.15 \\ \hline
\end{tabular}
\caption{Content selection performance in terms caputred propositions, 
for heuristics using memory tree sizes of $20$, computed on the validation set.
%for recall (R), precision (P), and F1-score 
%are given in percentages (validation set).
}
\label{table:cnt-sel}
\end{table}

\section{Conclusions}

We considered the problem of content selection in unsupervised extractive summarization, experimenting with scientific articles in the biomedical domain.
We explored a wide variety of heuristics that exploit properties of tree structures of content units as modeled by a psycho-linguistic model of reading comprehension, KvD \cite{kintsch1978toward}.
Results showed that heuristics leveraging tree properties
%, such as node depth and size of subtree, 
perform better than heuristics using plain frequency counts, a conclusion that holds when analyzing the selected content units.
%When analyzed at the content unit level, we find that tree-aware heuristics are capable of capturing more summary-worthy content units compared with frequency heuristics.
Inspecting the output of our systems in more detail, we noticed that they tend to extract sentences in close vicinity of each other.
This behaviour can be explained by the tendency of memory trees to retain propositions reflecting the topic being discussed at that point during the reading simulation.

Additionally, we argue about the necessity of comparing against a supervised baseline, specially if the proposed approach needs gold data to be fine-tuned on.
When comparing a recent unsupervised approach against a supervised baseline trained on the same small fine-tuning data, the supervised model outperforms all unsupervised configurations.
%Results of this kind cast shadows of doubt in the practical usability of a model.
%We encourage the NLP community to always include a supervised approach for reference, not only as a performance upper-bound but also to validate the applicability of a model.

\bibliography{draft}

\begin{thebibliography}{20}
\expandafter\ifx\csname natexlab\endcsname\relax\def\natexlab#1{#1}\fi

\bibitem[{Beltagy et~al.(2019)Beltagy, Lo, and
  Cohan}]{beltagy-etal-2019-scibert}
Iz~Beltagy, Kyle Lo, and Arman Cohan. 2019.
\newblock \href {https://www.aclweb.org/anthology/D19-1371} {Scibert: A
  pretrained language model for scientific text}.
\newblock In \emph{EMNLP}. Association for Computational Linguistics.

\bibitem[{Clarke and Lapata(2010)}]{clarke2010discourse}
James Clarke and Mirella Lapata. 2010.
\newblock Discourse constraints for document compression.
\newblock \emph{Computational Linguistics}, 36(3):411--441.

\bibitem[{Cohan et~al.(2019)Cohan, Beltagy, King, Dalvi, and
  Weld}]{cohan2019pretrained}
Arman Cohan, Iz~Beltagy, Daniel King, Bhavana Dalvi, and Daniel~S Weld. 2019.
\newblock Pretrained language models for sequential sentence classification.
\newblock In \emph{Proceedings of the 2019 Conference on Empirical Methods in
  Natural Language Processing and the 9th International Joint Conference on
  Natural Language Processing (EMNLP-IJCNLP)}, pages 3684--3690.

\bibitem[{Cohan et~al.(2018)Cohan, Dernoncourt, Kim, Bui, Kim, Chang, and
  Goharian}]{cohan2018discourse}
Arman Cohan, Franck Dernoncourt, Doo~Soon Kim, Trung Bui, Seokhwan Kim, Walter
  Chang, and Nazli Goharian. 2018.
\newblock A discourse-aware attention model for abstractive summarization of
  long documents.
\newblock \emph{arXiv preprint arXiv:1804.05685}.

\bibitem[{Fang(2019)}]{fang2019proposition}
Yimai Fang. 2019.
\newblock \emph{Proposition-based summarization with a coherence-driven
  incremental model}.
\newblock Ph.D. thesis, University of Cambridge.

\bibitem[{Fang and Teufel(2014)}]{fang2014summariser}
Yimai Fang and Simone Teufel. 2014.
\newblock A summariser based on human memory limitations and lexical
  competition.
\newblock In \emph{Proceedings of the 14th Conference of the European Chapter
  of the Association for Computational Linguistics}, pages 732--741.

\bibitem[{Hermann et~al.(2015)Hermann, Kocisky, Grefenstette, Espeholt, Kay,
  Suleyman, and Blunsom}]{hermann2015teaching}
Karl~Moritz Hermann, Tomas Kocisky, Edward Grefenstette, Lasse Espeholt, Will
  Kay, Mustafa Suleyman, and Phil Blunsom. 2015.
\newblock Teaching machines to read and comprehend.
\newblock In \emph{Advances in neural information processing systems}, pages
  1693--1701.

\bibitem[{Kedzie et~al.(2018)Kedzie, McKeown, and
  Daum{\'e}~III}]{kedzie2018content}
Chris Kedzie, Kathleen McKeown, and Hal Daum{\'e}~III. 2018.
\newblock Content selection in deep learning models of summarization.
\newblock In \emph{Proceedings of the 2018 Conference on Empirical Methods in
  Natural Language Processing}, pages 1818--1828.

\bibitem[{Kintsch and van Dijk(1978)}]{kintsch1978toward}
Walter Kintsch and Teun~A van Dijk. 1978.
\newblock Toward a model of text comprehension and production.
\newblock \emph{Psychological review}, 85(5):363.

\bibitem[{Lin(2004)}]{lin2004rouge}
Chin-Yew Lin. 2004.
\newblock Rouge: A package for automatic evaluation of summaries.
\newblock In \emph{Text summarization branches out}, pages 74--81.

\bibitem[{Lloret(2012)}]{lloret2012text}
Elena Lloret. 2012.
\newblock Text summarisation based on human language technologies and its
  applications.
\newblock \emph{Procesamiento del lenguaje natural}, (48):119--122.

\bibitem[{Loshchilov and Hutter(2018)}]{loshchilov2018decoupled}
Ilya Loshchilov and Frank Hutter. 2018.
\newblock Decoupled weight decay regularization.
\newblock In \emph{International Conference on Learning Representations}.

\bibitem[{Manning et~al.(2014)Manning, Surdeanu, Bauer, Finkel, Bethard, and
  McClosky}]{stanfordcorenlp}
Christopher~D. Manning, Mihai Surdeanu, John Bauer, Jenny Finkel, Steven~J.
  Bethard, and David McClosky. 2014.
\newblock \href {http://www.aclweb.org/anthology/P/P14/P14-5010} {The
  {Stanford} {CoreNLP} natural language processing toolkit}.
\newblock In \emph{Association for Computational Linguistics (ACL) System
  Demonstrations}, pages 55--60.

\bibitem[{Mihalcea and Tarau(2004)}]{mihalcea2004textrank}
Rada Mihalcea and Paul Tarau. 2004.
\newblock Textrank: Bringing order into text.
\newblock In \emph{Proceedings of the 2004 conference on empirical methods in
  natural language processing}, pages 404--411.

\bibitem[{Narayan et~al.(2018{\natexlab{a}})Narayan, Cardenas,
  Papasarantopoulos, Cohen, Lapata, Yu, and Chang}]{narayan2018document}
Shashi Narayan, Ronald Cardenas, Nikos Papasarantopoulos, Shay~B Cohen, Mirella
  Lapata, Jiangsheng Yu, and Yi~Chang. 2018{\natexlab{a}}.
\newblock Document modeling with external attention for sentence extraction.
\newblock In \emph{Proceedings of the 56th Annual Meeting of the Association
  for Computational Linguistics (Volume 1: Long Papers)}, pages 2020--2030.

\bibitem[{Narayan et~al.(2018{\natexlab{b}})Narayan, Cohen, and
  Lapata}]{narayan2018ranking}
Shashi Narayan, Shay~B Cohen, and Mirella Lapata. 2018{\natexlab{b}}.
\newblock Ranking sentences for extractive summarization with reinforcement
  learning.
\newblock In \emph{Proceedings of the 2018 Conference of the North American
  Chapter of the Association for Computational Linguistics: Human Language
  Technologies, Volume 1 (Long Papers)}, pages 1747--1759.

\bibitem[{Narayan et~al.(2019)Narayan, Cohen, and Lapata}]{narayan2019article}
Shashi Narayan, Shay~B Cohen, and Mirella Lapata. 2019.
\newblock What is this article about? extreme summarization with topic-aware
  convolutional neural networks.
\newblock \emph{Journal of Artificial Intelligence Research}, 66:243--278.

\bibitem[{Schumann et~al.(2020)Schumann, Mou, Lu, Vechtomova, and
  Markert}]{schumann2020discrete}
Raphael Schumann, Lili Mou, Yao Lu, Olga Vechtomova, and Katja Markert. 2020.
\newblock \href {https://doi.org/10.18653/v1/2020.acl-main.452} {Discrete
  optimization for unsupervised sentence summarization with word-level
  extraction}.
\newblock In \emph{Proceedings of the 58th Annual Meeting of the Association
  for Computational Linguistics}, pages 5032--5042, Online. Association for
  Computational Linguistics.

\bibitem[{Zhang et~al.(2016)Zhang, Li, Liu, and Gao}]{zhang2016coherent}
Renxian Zhang, Wenjie Li, Naishi Liu, and Dehong Gao. 2016.
\newblock Coherent narrative summarization with a cognitive model.
\newblock \emph{Computer Speech \& Language}, 35:134--160.

\bibitem[{Zheng and Lapata(2019)}]{zheng2019pacsum}
Hao Zheng and Mirella Lapata. 2019.
\newblock Sentence centrality revisited for unsupervised summarization.
\newblock In \emph{Proceedings of the 57th Annual Meeting of the Association
  for Computational Linguistics}, pages 6236--6247.

\end{thebibliography}
\bibliographystyle{acl_natbib}

\end{document}

\maketitle
\
\appendix

\section{Training Details \texttt{SciBert}}
\label{app:scibert}
The fine-tuning is trained for two epochs using a batch size of 8, document chunk size of 5.
For optimization, we use Adam optimizer \cite{loshchilov2018decoupled} with a fixed weight decay parameter of $0.1$.
Additionally, we use a slanted triangular learning rate scheduling with $10\%$ of total training steps as warm up and a top value of $1e^{-5}$.
We accumulate gradients for 16 training steps and clip gradients by norm value at $0.1$.

\section{Example Output}

\begin{table*}[]
\centering
\footnotesize
\begin{tabular}{|r|p{13cm}|l|}
\hline
ID & System Output                                                                                                                                                                                                                                                                                                       & Score  \\ \hline
   & \textbf{Reference} ($|\mathcal{S}|=54$)                                                                                                                                                                                                                                                                                                  &        \\
   & chorea is a neurological adverse effect of oral contraceptive pills ( ocps ) .                                                                                                                                                                                                                                      &        \\
   & the onset of chorea following ocps usage varies widely from few weeks to several years .                                                                                                                                                                                                                            &        \\
   & we report a rare case of chorea which developed within a week of starting ocps in an adolescent girl with polycystic ovarian disease .                                                                                                                                                                              &        \\ \hline
   & \textbf{Oracle} ($|\mathcal{S}|=213$)                                                                                                                                                                                                                                                                                                    &        \\
0  & oral contraceptive pills ( ocps ) are commonly used for contraception , polycystic ovarian syndrome , amenorrhea , menorrhagia , dysmenorrhea , endometriosis , etc .                                                                                                                                               &        \\
2  & neurological adverse effects associated with ocps are migraine , depression , psychosis , and cerebral infarction .                                                                                                                                                                                                 &        \\
4  & the onset of chorea usually varies from few weeks to several years after starting therapy . here                                                                                                                                                                                                                    &        \\
5  & , we report a case of chorea developing within 1 week of initiating ocps in an adolescent girl with polycystic ovarian disease .                                                                                                                                                                                    &        \\
7  & chorea is an abnormal involuntary movement disorder characterized by brief , semi - directed , irregular movements that are not repetitive or rhythmic , but appear to flow from one muscle to the next .                                                                                                           &        \\
12 & the onset of chorea varies widely from few weeks to several years ( average duration : 9 weeks ) .                                                                                                                                                                                                                  &        \\
13 & the shortest time interval between ocp intake and development of chorea reported in the literature is eight weeks in the patient with no significant medical history . in most cases , symptoms subside promptly after withdrawing the offending drug , usually within a period of 23 months . in this index case , &        \\
15 & prior knowledge of this rare adverse effect and high index of suspicion can help in early diagnosis and discontinuation of therapy .                                                                                                                                                                                &        \\ \hline
   & \textbf{Heuristic }($|\mathcal{S}|=205$)                                                                                                                                                                                                                                                                                                 &        \\
0  & oral contraceptive pills ( ocps ) are commonly used for contraception , polycystic ovarian syndrome , amenorrhea , menorrhagia , dysmenorrhea , endometriosis , etc .                                                                                                                                               & 1.9409 \\
1  & , common side effects of ocps include nausea , vomiting , headache , bloating , breast tenderness , swelling of the feet , weight gain , breakthrough bleeding , and venous thromboembolism .                                                                                                                       & 6.7545 \\
2  & neurological adverse effects associated with ocps are migraine , depression , psychosis , and cerebral infarction .                                                                                                                                                                                                 & 2.0535 \\
3  & chorea caused by ocps is extremely rare and only very few cases are reported in the literature .                                                                                                                                                                                                                    & 1.2707 \\
4  & the onset of chorea usually varies from few weeks to several years after starting therapy . here                                                                                                                                                                                                                    & 1.5047 \\
5  & , we report a case of chorea developing within 1 week of initiating ocps in an adolescent girl with polycystic ovarian disease .                                                                                                                                                                                    & 2.3248 \\
6  & neurological side effects of ocps have been linked to alterations in coagulation pathways and secondary vascular complications .                                                                                                                                                                                    & 5.0742 \\
8  & the first description of chorea associated with ocps was reported by fernando in 1966 and subsequently few similar published reports established ocps as a cause of chorea .                                                                                                                                        & 6.7902 \\
9  & no particular oral contraceptive preparation was consistently associated with chorea , and both low and high - dose preparations can cause this movement disorder .                                                                                                                                                 & 2.9526 \\ \hline
   & \textbf{Top 5 propositions }  &  \\
   & associated(the\_description; with\_ocps)(0.9769)                                                                                                                                                                                                                                                                    &        \\
   & have\_been\_linked(effects; to\_alterations)(0.8413)                                                                                                                                                                                                                                                                &        \\
   & of(side; ocps)(0.8201)                                                                                                                                                                                                                                                                                              &        \\
   & of(the\_description; chorea)(0.7976)                                                                                                                                                                                                                                                                                &        \\
   & of(effects; ocps)(0.7815)                                                                                                                                                                                                                                                                                           &       \\ \hline
\end{tabular}
\label{table:output-exp}
\caption{Example of reference summary (abstract of the article; top row), summary extracted by the oracle (second row), and summary produced by our best heuristic (third row), \textsc{Sub-exp-rprob} for a tree size of $M=20$. Bottom row features the top 5 propositions scored by the heuristic.}
\end{table*}